# NLDS-UCSC at SemEval-2016 Task 6: A Semi-Supervised Approach to Detecting Stance in Tweets

**Amita Misra, Brian Ecker, Theodore Handleman, Nicolas Hahn** and **Marilyn Walker**
Natural Language and Dialogue Systems Lab
University of California, Santa Cruz
1156 N. High. SOE-3
Santa Cruz, California, 95064, USA
amisra2|becker|thandlem|nhahn|mawalker@ucsc.edu

## Abstract

Stance classification aims to identify, for a particular issue under discussion, whether the speaker or author of a conversational turn has Pro (Favor) or Con (Against) stance on the issue. Detecting stance in tweets is a new task proposed for SemEval-2016 Task6, involving predicting stance for a dataset of tweets on the topics of abortion, atheism, climate change, feminism and Hillary Clinton. Given the small size of the dataset, our team created our own topic-specific training corpus by developing a set of high precision hashtags for each topic that were used to query the twitter API, with the aim of developing a large training corpus without additional human labeling of tweets for stance. The hashtags selected for each topic were predicted to be stance-bearing on their own. Experimental results demonstrate good performance for our features for opinion-target pairs based on generalizing dependency features using sentiment lexicons.

## 1 Introduction

Social media websites such as microblogs, weblogs, and discussion forums are used by millions of users to express their opinions on almost everything from brands, celebrities, and events to important social and political issues. In recent years, the microblogging service Twitter has emerged as one of the most popular and useful sources of user content, and recent research has begun to develop tools and computational models for tweet-level opinion and sentiment analysis. Stance classification aims to identify, for a particular issue under discussion, whether the speaker or author of a conversational turn has a Pro (Favor) or Con (Against) stance on the issue (Somasundaran and Wiebe, 2009; Somasundaran and Wiebe, 2010; Walker et al., 2012c; Sridhar et al., 2015; Hasan and Ng, 2013).

| Id | Tweet | Stance |
|---|---|---|
| T1 | We are causing the ice masses of Earth to melt at an alarming rate. | FAVOR |
| T2 | ONE Volcano emits more pollution than man has in our HISTORY! | AGAINST |
| T3 | It's most exciting to witness a major development! | NONE |
| T4 | The Weather app keeps taunting us with rain. #PNW #drought | NONE |

Table 1: Example tweets with stance labels for the issue *Climate Change is a Real Concern.*

Detecting stance in tweets is a new task proposed for SemEval-2016 (Mohammad et al., 2016). The aim of the task is to determine user stance (FAVOR, AGAINST, or NONE) in a dataset of tweets on the five selected topics of abortion, atheism, climate change, feminism and Hillary Clinton. Consider the tweets in Table 1, which express stance toward the target issue *Climate Change is a Real Concern.* It can be inferred that the author of tweet **T1** is in favor of the target while the author of tweet **T2** is clearly against the target. However due to the brevity of tweets, there is not always sufficient information about the target to determine stance: in the case of tweet **T3**, we are unsure what *major development* the user is talking about. In the case of tweet **T4**, we know the user acknowledges the existence of a drought, but we do not know their stance on the issue of climate change solely based on this information. In such cases the stance of the tweets is labelled NONE for this issue.

The task is nontrivial due to the challenges of the tweet genre. Tweets are often highly informal with language that is colorful and ungrammatical. They may also involve sarcasm, making opinion-mining tasks more challenging (Riloff et al., 2013; Reyes et al., 2012). Users may assert their stance using factual or emotional content, and due to their restricted length, tweets may not be well structured or coherent. As a result, NLP tools trained on well-structured text do not work well in Twitter (Dey and Haque, 2008), and new tools are constantly being developed (Qadir and Riloff, 2014; Kong et al., 2014; Han and Baldwin, 2011; Zhu et al., 2014).

Our approach to stance classification in tweets is primarily based on developing a suite of tools for processing Twitter that mirrors our previous work on stance classification in online forums (Walker et al., 2012c; Sridhar et al., 2015; Anand et al., 2011; Walker et al., 2012b; Misra and Walker, 2015). We develop generalized dependency features that capture expressed sentiment or attitude towards particular targets, using the Tweebo dependency parser (Kong et al., 2014). Given the small size of the official task dataset, we created our own topic-specific training corpus in a semi-supervised manner. We developed a set of high precision hashtags for each topic that were used to query the Twitter API in order to create a large training corpus without additional human labeling of tweets for stance. The hashtags and boolean combinations of hashatgs selected for each topic were predicted to be stance-bearing on their own. See Table 2.

There has been considerable previous work on stance classification in online forums and in congressional debates (Thomas et al., 2006; Burfoot et al., 2011; Somasundaran and Wiebe, 2009; Somasundaran and Wiebe, 2010; Walker et al., 2012c; Sridhar et al., 2015; Hasan and Ng, 2013; Boltuzic and Šnajder, 2014; Hasan and Ng, 2014). A number of these studies show that collective classification approaches perform well, and that the context (Walker et al., 2012c; Abbott et al., 2011), and meta information such as author constraints are useful for stance classification (Hassan et al., 2012; Hasan and Ng, 2014). Collective classification is not possible in the current task because the only information provided is the text of each individual tweet. Inspired by earlier work (Joshi and Penstein-Rosé, 2009; Somasundaran and Wiebe, 2009; Somasundaran and Wiebe, 2010; Walker et al., 2012b), we apply a framework for developing features for opinion-target pairs based on generalized structural dependency features, using the LIWC dictionary as the basis for generalization (Pennebaker et al., 2001). We also develop features to capture domain knowledge using PMI values for topic n-grams in order to improve the recognition of tweets with the NONE stance. We describe our system and data in Sec. 2, our experimental set-up in Sec. 3, and our results and error analysis in Sec. 4. We conclude and discuss future directions in Sec. 5.

## 2 Data

The relatively small, unbalanced training set provided for the task introduced an interesting subtask for precise topic-oriented tweet collection without direct human annotation. Twitter hashtags provide a method for users to tag their own content by topic, and we exploit this self-annotation to collect a larger dataset for training by hand-selecting seed hashtags for both the FAVOR and AGAINST stances for each topic. We then query Twitter for tweets containing these hashtags using the API, and produce a training set from the results without further supervision. We treat the original SemEval dataset as development data, assuming it is similar to the SemEval test set.

When collecting data in this fashion, there are multiple factors that must be accounted for. These include the accuracy of labels, data uniformity and representativeness, and dataset size. The accuracy of our labels is directly related to the specificity of our hashtags. We perform a small evaluation of each hashtag added to our seed pool by checking it's accuracy on a subset of queried data by hand. With regard to data uniformity and representativeness, we want to ensure that our collected data is not too uniform as a large collection of very similar tweets provides little additional information, and we want to ensure that our data is representative of the actual SemEval data that may be produced from different preprocessing and collection techniques. We evaluate the uniformity and representativeness of our data in Sec. 4.

After we finish collecting data, we create bal-anced FAVOR, AGAINST, and NONE training sets for each topic. Tweets in the NONE class are collected from other topics or from a corpus of random tweets. A summary of the final seed hashtags and dataset sizes is shown in Table 2.

| Topic | Seed Hashtags | | Tweets |
|---|---|---|---|
| | FAVOR | AGAINST | #N |
| Abortion | #ProChoice, #StandWithPP, #IStandWithPP, #RightToChoose | #ProLife, #UnbornLivesMatter, #DefundPP, #PrayToEndAbortion | 22194 |
| Atheism | #Awesome∧atheism, #AtheistVoter, #AntiReligion, #NotAfraidOfBurningInHell | #AntiAtheism, #HolyBible, #God, #PrayerWorks, #PraiseTheLord | 11616 |
| Climate Change | #DemandClimateAction, #SaveThePlanet, #ActOnClimate, #GlobalWarmingIsReal | #Lies∧climate, #Hoax∧climate, #GlobalWarmingIsALie, #Fraud∧climate | 321 |
| Feminism | #YayFeminism, #YesAllWomen ∧ feminism, #FeministsAreBeautiful | #AntiFeminism, #AntiFeminist, #WomenAgainstFeminism | 4446 |
| Hillary Clinton | #ImWithHer, #HillYes, #ITrustHillary, #TeamHillary | #StopHillary, #OhHillNo, #HillNo, #HillaryForPrison, #WhyImNotVotingForHillary | 8529 |

Table 2: Example hashtags and hashtag boolean combinations used to produce training data, and size of the resulting final balanced training dataset.

## 2.1 Data preprocessing

Since tweets can be noisy, uninformative, or ambiguously labeled, we apply the three filters below to get better quality tweets.

- *Duplicate removal*: Remove all tweets that have an 80% or greater overlap with another already included tweet.

- *Dictionary words*: Tweets with less than 4 dictionary words are excluded. Although this filter may not be appropriate for all tasks as Tweets may incorporate large amounts of non-dictionary slang, we observe that the SemEval training data has few instances that do not pass this test.

- *Favor and Against*: Remove tweets that have both FAVOR and AGAINST hashtags.

## 2.2 Data Normalization

Tweets can be noisy due to irregular words and other genre specific language. We preprocess all tweets as follows:

- *Repeated characters*: Replace a sequence of repeated characters by two characters. For example, convert "shooooooooot" to "shoot".

- *Lexical variation*: We used the Python Enchant dictionary to determine if a token is a dictionary word. If a token is not present in the dictionary then it is replaced by finding a possible lexical variant using the English Social Media Normalisation Lexicon (Han et al., 2012), for example, *"tmrrw"* is changed to *"tomorrow"*.

We use a part-of-speech tagger for tweets to perform tokenization and POS labelling (Gimpel et al., 2011). We also use TweeboParser, a dependency parser specifically designed for tweets, to parse each tweet (Kong et al., 2014).

Our data representation for the corpus keeps track of the original tweet, the normalization replacements, the POS tags, and the parses. Sec. 3 describes the features derived from these pre-processing steps that we also store in our corpus database, modelled after IAC 2.0. (Abbott et al., 2016).

## 3 Experimental Setup

We explored a large number of machine learning algorithms and feature combinations, using the automatically harvested tweets as training and the training set provided for the task as our development data to fit the parameters for the final submitted NLDS-UCSC system. Sec. 3.1 describes the feature sets created using the development set. To evaluate the effects of hashtags on the test set we explored two different ways to train the system. Table 4 presents the results on the test set with hashtags present in the dataset while Table 5 is the performance without hashtags.

| Topic | Features | Feature Selection | F_score (evaluation metric ) | | |
|---|---|---|---|---|---|
| | | | favor | against | average |
| **Abortion** | unigram, bigram, dep, liwc_dep, opinion_dep | none | 0.71 | 0.64 | 0.67 |
| **Atheism** | unigram, bigram, dep, liwc_dep, opinion_dep | correlation | 0.73 | 0.66 | 0.69 |
| **Climate Change** | unigram, bigram, liwc_dep, opinion_dep, POS_bigram, POS_trigram, LIWC, high_pmi_n-gram_count, max_pmi, high_pmi_in_topic | gainratio | 0.53 | 0.67 | 0.60 |
| **Feminism** | unigram, bigram, dep, liwc_dep, opinion_dep, POS_bigram | correlation | 0.55 | 0.57 | 0.56 |
| **Hillary Clinton** | unigram | none | 0.63 | 0.60 | 0.61 |

Table 3: Best performing model for each topic on Dev Set w/ hashtags, along with F-measure for favor, against, and their average.

| Features | Abortion | | | Atheism | | | Climate Change | | | Feminism | | | Hillary Clinton | | |
|---|---|---|---|---|---|---|---|---|---|---|---|---|---|---|---|
| | favor | against | avg | favor | against | avg | favor | against | avg | favor | against | avg | favor | against | avg |
| **Unigram** | 0.55 | 0.72 | 0.63 | 0.51 | 0.77 | 0.64 | 0.61 | 0.22 | 0.41 | 0.33 | 0.64 | 0.49 | 0.42 | 0.73 | 0.57 |
| **All dependencies** | 0.48 | 0.68 | 0.58 | 0.41 | 0.76 | 0.59 | 0.55 | 0.16 | 0.36 | 0.35 | 0.61 | 0.48 | 0.40 | 0.66 | 0.52 |
| **POS n-gram** | 0.40 | 0.47 | 0.44 | 0.36 | 0.75 | 0.55 | 0.58 | 0.11 | 0.35 | 0.26 | 0.47 | 0.37 | 0.31 | 0.52 | 0.42 |
| **LIWC** | 0.42 | 0.44 | 0.43 | 0.36 | 0.66 | 0.51 | 0.59 | 0.18 | 0.38 | 0.28 | 0.49 | 0.39 | 0.28 | 0.42 | 0.35 |
| **PMI** | 0.40 | 0.65 | 0.52 | 0.32 | 0.58 | 0.45 | 0.52 | 0.12 | 0.32 | 0.30 | 0.49 | 0.39 | 0.1 | 0.49 | 0.30 |
| **Best model (Dev)** | 0.52 | 0.73 | 0.62 | 0.43 | 0.72 | 0.57 | 0.69 | 0.15 | 0.42 | 0.36 | 0.64 | 0.50 | 0.42 | 0.73 | 0.57 |

Table 4: Feature ablation w/ hashtags for each topic on Test Set, along with F-measure for favor, against, and their average.

### 3.1 Features

**Unigrams and Bigrams:** We extracted unigrams and bigrams from the preprocessed tweets. The useful unigrams are mainly hashtags. We used both stemmed and unstemmed ngrams.

**POS bigrams and trigrams:** The tweet part-of-speech tagger is used to perform tokenization and POS identification (Gimpel et al., 2011). We then extracted POS bigrams and trigrams as features.

**LIWC:** We derived features using the Linguistics Inquiry Word Count tool and use the count of words in each category as the feature value (Pennebaker et al., 2001).

**Dependency:** We used TweeboParser to extract dependency features. For a given tweet, TweeboParser predicts its syntactic structure, represented by unlabeled dependencies (Kong et al., 2014).

**Generalized LIWC and Opinion Dependency:** We created two kinds of generalized dependency features. Building on the idea that partially generalized dependencies are better than ungeneralized or completely generalized dependencies (Joshi and Penstein-Rosé, 2009), we leave one dependency element lexicalized and generalize the other to its LIWC category for LIWC dependency features. We follow a similar process to produce generalized opinion dependencies using AFINN lexicon and opinion-lexicon-English by (Hu and Liu) replacing one element of the dependency with its sentiment score and leaving the other element lexicalized (Hu and Liu, 2004; Nielsen, 2011). [1] [2]

Inspired by previous work on combining sentiment lexicons we used a combined sentiment score to denote the accuracy of a sentiment word rather than its strength. If dictionaries contradict one another on the sentiment polarity for a word, then the score is neutralized to zero. If a single dictionary lists the polarity word, but it is unlisted or neutral in the other dictionary, then the score is 1 in the direction of the polarity. If both dictionaries list a word with the same polarity, then the score is 2 in the direction of the polarity. After calculating the combined sentiment score, we check if either of the previous two words is listed as a negation by LIWC,

---
[1] https://www.cs.uic.edu/ liub/FBS /sentiment-analysis.html
[2] http://neuro.imm.dtu.dk/wiki/AFINN

and invert the polarity if a negation is found(Cho et al., 2013; Hasan and Ng, 2012).

**Pointwise Mutual Information (PMI):** For each topic, we calculate normalized pointwise mutual information over a combination of an extended version of IAC 2.0, a topic annotated database of posts from debate forums (Abbott et al., 2016; Walker et al., 2012a), and our own collected tweets for each topic. IAC 2.0 includes several topics that are in overlap with the topics in the current task.

We then create a pool of top-N percent PMI unigrams, bigrams, and trigrams for each topic and use the count of words in each tweet that are also in this pool as a feature. We also use the highest PMI value of an n-gram in each tweet as a feature.

## 4 Results

We ran experiments using the SemEval training as our development data with NaiveBayesMultinomial, SVM, and J48 from WEKA. We tried a large number of feature combinations w/ and w/out stemmed n-grams.

The best performing system ended up being different for each topic. Overall, NBM worked the best for all topics. Table 3 describes the system model submitted for each topic based on the results. The model that performed best on the dev set was used to report accuracies on the test set. We present the results on the test set in Table 4. The best performing model from the dev set for the topic of climate control shows a marginal but not significant improvement, and none of the features on their own could beat a unigram baseline for any topic. Since most of the hashatgs are unigrams, we hypothesize this may be due to presence of strong stance-bearing hashtags in the data. In order to assess the issue of strong stance-bearing hashtags still existing in the training data, we remove all hashtags from both the training data and test data and retrain classifiers. See Table 5. For the topic of abortion we see that the combined unigram, bigram, and dependency model now outperforms the unigram model, and the dependencies alone start to edge out an advantage as well, suggesting that it may be the case that strong stance-bearing hashtags distract from and disguise true performance of each feature and potentially the model as a whole.

Feature ablation results reveal that for the majority of the topics part-of-speech n-grams perform better than LIWC. This was surprising because LIWC was designed to capture emotional and psychological behavior in conversations, and because previous research on stance classification using debate forums shows LIWC categories can improve an n-gram baseline (Anand et al., 2011). This may be due to *sarcam and irony*, a frequent phenomena in twitter not captured by LIWC, but which may to some extent be captured by part-of-speech n-grams that reflect the use of adjectives and adverbs in sarcastic posts (Lukin and Walker, 2013; Reyes et al., 2012). Generalizing Twitter-specific dependency structures using LIWC and sentiment lexicons does however prove useful.

### 4.1 Learning Curves

To asses the usefulness of increasing our training set size we plot learning curves for each topic (abortion shown in Figure 1 and all others in appendix A). For each topic we plot the average f-measure (FAVOR, AGAINST) for a unigram baseline, dependency baseline, and best performing model on the dev set.

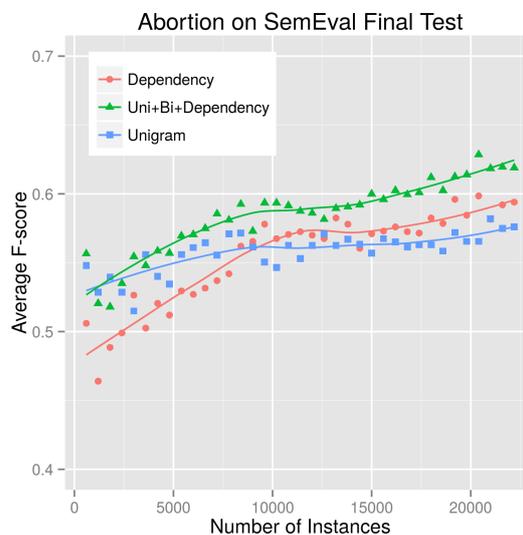

Figure 1: Training set size vs average Fscore for abortion.

Figure 1 shows that the classifier for the abortion topic gains around 0.6 f-measure when increasing the number of instances from 5,000 to 20,000, and it continues to show promise for growth, especially in terms of the dependency and best model curves. We see a similar promise for growth in the model for Hillary Clinton in which dependencies have just passed unigrams. The differences in learning rate for each topic suggest that the precision of our stance-sided seed hashtags varies largely by topic because similar amounts of data provide less information gain, signaling the data may be of lower quality. In addition to extracting more data for each topic, it would also be helpful to refine our hashtag selection for topics such as atheism, where increase in training set size do not yield performance improvement.

| Features | Abortion | | | Atheism | | | Climate Change | | | Feminism | | | Hillary Clinton | | |
|---|---|---|---|---|---|---|---|---|---|---|---|---|---|---|---|
| | favor | against | avg | favor | against | avg | favor | against | avg | favor | against | avg | favor | against | avg |
| unigram | 0.50 | 0.70 | 0.60 | 0.49 | 0.77 | 0.63 | 0.62 | 0.20 | 0.41 | 0.32 | 0.63 | 0.47 | 0.35 | 0.67 | 0.51 |
| All dependencies | 0.50 | 0.71 | 0.61 | 0.41 | 0.77 | 0.59 | 0.63 | 0.12 | 0.38 | 0.36 | 0.62 | 0.49 | 0.40 | 0.68 | 0.54 |
| POS | 0.39 | 0.44 | 0.42 | 0.29 | 0.70 | 0.50 | 0.62 | 0.09 | 0.35 | 0.30 | 0.48 | 0.39 | 0.32 | 0.5 | 0.41 |
| POS+Dep | .51 | 0.73 | 0.62 | 0.47 | 0.81 | 0.64 | 0.66 | .09 | 0.38 | 0.32 | 0.61 | 0.46 | 0.42 | 0.69 | 0.56 |

Table 5: Tweets without hashtags on the Test set, along with F-measure for favor, against and their average.

## 5 Conclusion and Future work

We explore a semi-supervised approach to stance classification using stance-bearing hashtags and achieve reasonable accuracies on a hand-annotated test set. This suggests that our approach of querying using seed hashtags and using some heuristic filters to improve tweet quality may be promising for generating a large corpus of training data. It may also be useful in other domains where hand-annotated data does not exist and getting annotations is time consuming and costly effort. To determine the feasibility of using this semi-supervised data in other domains, we removed all the hashtags from the tweets and again compared the performance of our dependency and unigram features. Table 5 shows that these results look promising. In future work, we hope to use more intelligent features that may capture irony and sarcasm, and we plan to expand and refine our data collection process to account for the varying precision of stance-sided hashtags across topics.

## 6 Acknowledgments

We are thankful to Yanfei Tu for her help on the project. This work was supported by the Robust Intelligence Program of NSF under Award #IIS-1302668.

## References

R Abbott, M Walker, J E. Fox Tree, Pranav Anand, Robeson Bowmani, and Joseph King. 2011. How can you say such things?!?: Recognizing Disagreement in Informal Political Argument. In *Proc of the ACL Workshop on Language and Social Media*.

R Abbott, B Ecker, P Anand, and M Walker. 2016. Internet argument corpus 2.0: An sql schema for dialogic social media and the corpora to go with it. In *Language Resources and Evaluation conf, LREC2016*.

P Anand, M Walker, R Abbott, J E. Fox Tree, Robeson Bowmani, and Michael Minor. 2011. Cats Rule and Dogs Drool: Classifying Stance in Online Debate. In *Proc of the ACL Workshop on Sentiment and Subjectivity*.

Filip Boltuzic and Jan Šnajder. 2014. Back up your stance: Recognizing arguments in online discussions. In *Proc of the First Workshop on Argumentation Mining*, pp 49–58.

Clinton Burfoot, Steven Bird, and Timothy Baldwin. 2011. Collective classification of congressional floor-debate transcripts. In *Association for Computational Linguistics (ACL)*, pp 1506–1515.

Heeryon Cho, Jong-Seok Lee, and Songkuk Kim. 2013. Enhancing lexicon-based review classification by merging and revising sentiment dictionaries. In *Sixth Inte Joint Conf on Natural Language Processing, IJCNLP 2013, 2013*, pp 463–470.

Lipika Dey and S K Mirajul Haque. 2008. Opinion mining from noisy text data. In *Proc of the Second Workshop on Analytics for Noisy Unstructured Text Data*, AND '08, pp 83–90, ACM.

Kevin Gimpel, Nathan Schneider, Brendan O'Connor, Dipanjan Das, Daniel Mills, Jacob Eisenstein, Michael Heilman, Dani Yogatama, Jeffrey Flanigan, and Noah A Smith. 2011. Part-of-speech tagging for twitter: Annotation, features, and experiments. In *Proc of the 49th Annual Meeting of the Assoc for Compu-* *tational Linguistics: Human Language Technologies: short papers-Volume 2*, pp 42–47.

Bo Han and Timothy Baldwin. 2011. Lexical normalisation of short text messages: Makn sens a# twitter. In *Proc of the 49th Annual Meeting of the Association for Computational Linguistics: Human Language Technologies-Volume 1*, pp 368–378. Association for Computational Linguistics.

Bo Han, Paul Cook, and Timothy Baldwin. 2012. Automatically constructing a normalisation dictionary for microblogs. In *Proc of the 2012 Joint conf on Empirical Methods in Natural Language Processing and Computational Natural Language Learning*, EMNLP-CoNLL '12, pp 421–432.

Kazi Saidul Hasan and Vincent Ng. 2012. Predicting stance in ideological debate with rich linguistic knowledge. In *Proc of COLING 2012: Posters*, pp 451–460.

Kazi Saidul Hasan and Vincent Ng. 2013. Stance classification of ideological debates: Data, models, features, and constraints. page 1348–1356.

Kazi Saidul Hasan and Vincent Ng. 2014. Why are you taking this stance? identifying and classifying reasons in ideological debates. In *Proc of the Conf on Empirical Methods in Natural Language Processing*.

Ahmed Hassan, Amjad Abu-Jbara, and Dragomir Radev. 2012. Detecting subgroups in online discussions by modeling positive and negative relations among participants. In *Proc of the 2012 Joint Conf on Empirical Methods in Natural Language Processing and Computational Natural Language Learning*.

Minqing Hu and Bing Liu. 2004. Mining and summarizing customer reviews. In *Proc of the Tenth ACM SIGKDD International conf on Knowledge Discovery and Data Mining*, KDD '04, pp 168–177.

M. Joshi and C. Penstein-Rosé. 2009. Generalizing dependency features for opinion mining. In *Proc of the ACL-IJCNLP 2009 conf Short Papers*, pp 313–316.

Lingpeng Kong, Nathan Schneider, Swabha Swayamdipta, Archna Bhatia, Chris Dyer, and Noah A Smith. 2014. A dependency parser for tweets. In *In Proc. of EMNLP*. Citeseer.

Stephanie Lukin and Marilyn Walker. 2013. Really? well, Apparently bootstrapping improves the performance of sarcasm and nastiness classifiers for online dialogue. *NAACL 2013*, page 30.

Amita Misra and Marilyn A Walker. 2015. Topic independent identification of agreement and disagreement in social media dialogue. In *Proc of the SIGDIAL 2013 conf: The 15th Annual Meeting of the Special Interest Group on Discourse and Dialogue*.

Saif M. Mohammad, Svetlana Kiritchenko, Parinaz Sobhani, Xiaodan Zhu, and Colin Cherry. 2016. Semeval-2016 task 6: Detecting stance in tweets. In *Proc of the Inte Workshop on Semantic Evaluation*, SemEval.

Finn Årup Nielsen. 2011. A new ANEW: evaluation of a word list for sentiment analysis in microblogs. In Matthew Rowe, Milan Stankovic, Aba-Sah Dadzie, and Mariann Hardey, editors, *Proc of the ESWC2011 Workshop on 'Making Sense of Microposts': Big things come in small packages*, 718 of *CEUR Workshop Proc*, pp 93–98.

J. W. Pennebaker, L. E. Francis, and R. J. Booth, 2001. *LIWC: Linguistic Inquiry and Word Count*.

Ashequl Qadir and Ellen Riloff. 2014. Learning emotion indicators from tweets: Hashtags, hashtag patterns, and phrases. In *Proc of the conf on Empirical Methods in Natural Language Processing (EMNLP), Association for Computational Linguistics*, pp 1203–1209.

A. Reyes, P. Rosso, and D. Buscaldi. 2012. From humor recognition to irony detection: The figurative language of social media. *Data & Knowledge Engineering*.

Ellen Riloff, Ashequl Qadir, Prafulla Surve, Lalindra De Silva, Nathan Gilbert, and Ruihong Huang. 2013. Sarcasm as contrast between a positive sentiment and negative situation. In *Proc of the 2013 conf on Empirical Methods in Natural Language Processing*.

S. Somasundaran and J. Wiebe. 2009. Recognizing stances in online debates. In *Proc of the 47th Annual Meeting of the ACL*, pp 226–234.

S. Somasundaran and J. Wiebe. 2010. Recognizing stances in ideological on-line debates. In *Proc of the NAACL HLT 2010 Workshop on Computational Approaches to Analysis and Generation of Emotion in Text*, pp 116–124.

Dhanya Sridhar, James Foulds, Bert Huang, Lise Getoor, and Marilyn Walker. 2015. Joint models of disagreement and stance in online debate. In *Annual Meeting of the Ass for Computational Linguistics (ACL)*.

M. Thomas, B. Pang, and L. Lee. 2006. Get out the vote: Determining support or opposition from Congressional floor-debate transcripts. In *Proc of the 2006 conf on empirical methods in natural language processing*, pp 327–335.

Marilyn Walker, Pranav Anand, Robert Abbott, and Jean E. Fox Tree. 2012a. A corpus for research on deliberation and debate. In *Language Resources and Evaluation conf, LREC2012*.

Marilyn A Walker, Pranav Anand, Rob Abbott, Jean E Fox Tree, Craig Martell, and Joseph King. 2012b. That is your evidence?: Classifying stance in online political debate. *Decision Support Systems*, 53(4):719–729.

Marilyn A Walker, Pranav Anand, Robert Abbott, and Ricky Grant. 2012c. Stance classification using dialogic properties of persuasion. In *Proc of the 2012* *conf of the North American Chapter of the Association for Computational Linguistics: Human Language Technologies*, pp 592–596.

Xiaodan Zhu, Svetlana Kiritchenko, and Saif M Mohammad. 2014. Nrc-canada-2014: Recent improvements in the sentiment analysis of tweets. In *Proc of the 8th Inter Workshop on Semantic Evaluation (SemEval 2014)*, pp 443–447.

## A Appendix: Learning Curves

Below we show the learning curves for each topic on the SemEval test data after removing hashtags from both the train and test sets. Climate change is excluded due to the small number of training instances. Each graph includes a line for unigrams, dependencies, and the best model on the dev set (unigrams are the best model for the Hillary Clinton topic).

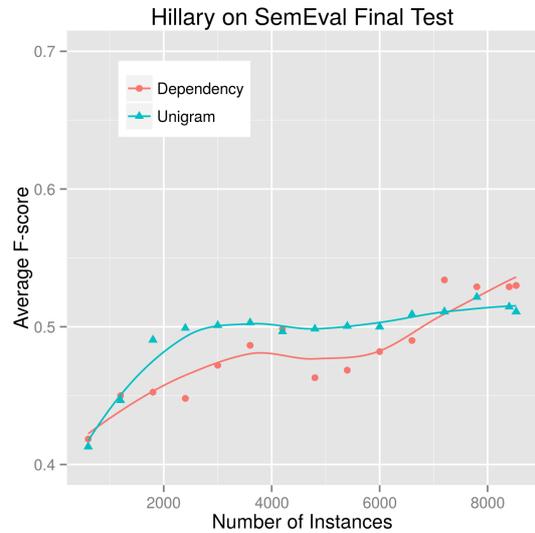

Figure 3: Training set size vs average Fscore for Hillary Clinton

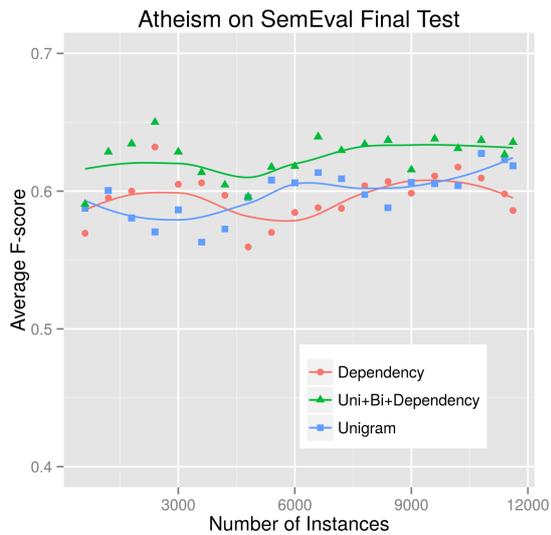

Figure 2: Training set size vs average Fscore for Atheism

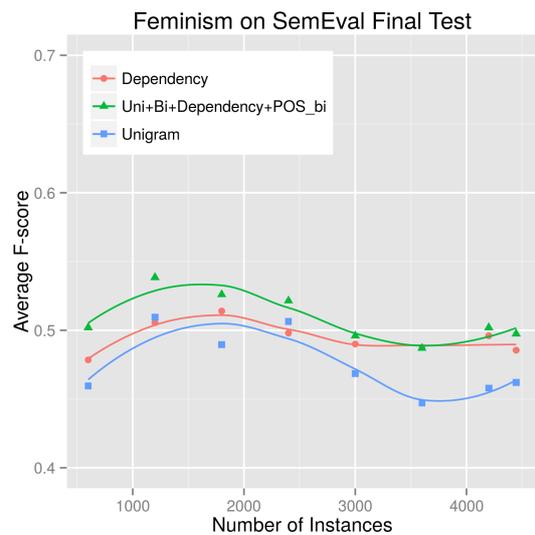

Figure 4: Training set size vs average Fscore for Feminism